\documentclass[10pt,twocolumn,letterpaper]{article}

\usepackage{wacv}
\usepackage{times}
\usepackage{epsfig}
\usepackage{graphicx}
\usepackage{amsmath}
\usepackage{amssymb}
\usepackage{listings}
\usepackage[titletoc,toc,title]{appendix}

\newcommand{\SotA} {state-of-the-art }

\usepackage[pagebackref=true,breaklinks=true,letterpaper=true,colorlinks,bookmarks=false]{hyperref}

\wacvfinalcopy 

\def\wacvPaperID{54} 

\ifwacvfinal\fi
\begin{document}

\title{Cyclical Learning Rates for Training Neural Networks}

\author{Leslie N.~Smith\\
U.S. Naval Research Laboratory, Code 5514\\
4555 Overlook Ave., SW., Washington, D.C.  20375\\
{\tt\small leslie.smith@nrl.navy.mil}
}

\maketitle
\ifwacvfinal\thispagestyle{empty}\fi

\begin{abstract}
It is known that the learning rate is the most important hyper-parameter to tune for training deep neural networks.
This paper describes a new method for setting the learning rate, named \textit{cyclical learning rates}, which practically eliminates the need to experimentally find the best values and schedule for the global learning rates. 
Instead of monotonically decreasing the learning rate, this method lets the learning rate cyclically vary between reasonable boundary values. 
Training with cyclical learning rates instead of fixed values achieves improved classification accuracy without a need to tune and often in fewer iterations.
This paper also describes a simple way to estimate ``reasonable bounds'' -- linearly increasing the learning rate of the network for a few epochs.
In addition, cyclical learning rates are demonstrated on the CIFAR-10 and CIFAR-100 datasets with ResNets, Stochastic Depth networks, and DenseNets, and the ImageNet dataset with the AlexNet and GoogLeNet architectures.
These are practical tools for everyone who trains neural networks.
\end{abstract}

\section{Introduction}
\label{sec:intro}

Deep neural networks are the basis of \SotA results for image recognition \cite{Alexnet12, simonyan2014very, szegedy2014going}, object detection \cite{girshick2014rich}, face recognition \cite{taigman2014deepface}, speech recognition \cite{graves2014towards}, machine translation \cite{sutskever2014sequence}, image caption generation \cite{vinyals2014show}, and driverless car technology \cite{huval2015empirical}.
However, training a deep neural network is a difficult global optimization problem.

A deep neural network is typically updated by stochastic gradient descent and the parameters $\theta$ (weights) are updated by
$\theta^t = \theta^{t-1} - \epsilon_t  \frac{\partial L}{\partial \theta}$,
where L is a loss function and $\epsilon_t$ is the learning rate.
It is well known that too small a learning rate will make a training algorithm converge slowly while too large a learning rate will make the training algorithm diverge \cite{Bengio12}.
Hence, one must experiment with a variety of learning rates and schedules.


\begin{figure}[t]
  \vspace{-10pt}
  \begin{center}
   \includegraphics[width=1.0\linewidth]{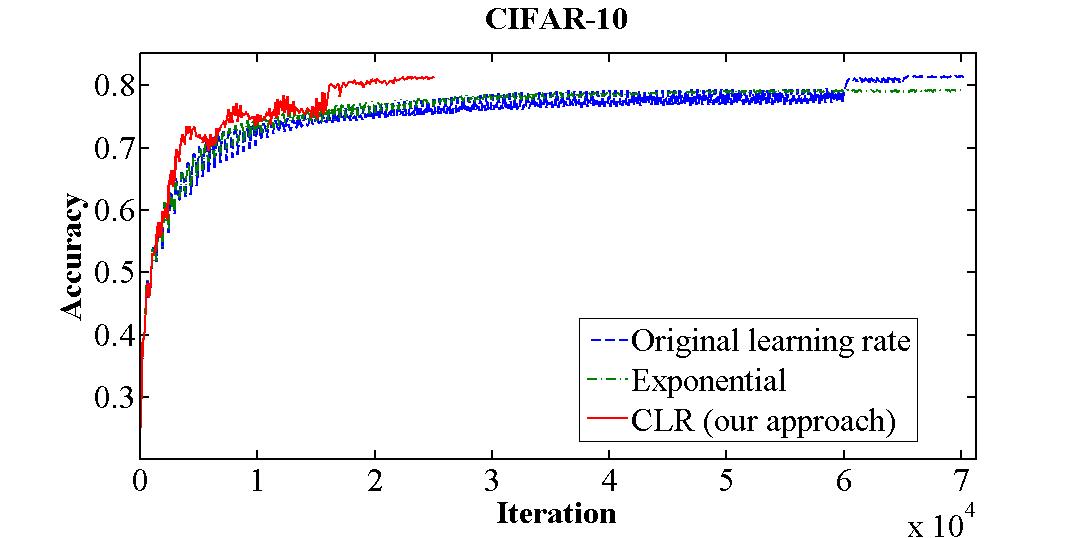}
  \end{center}
  \vspace{-5pt}
   \caption{Classification accuracy while training CIFAR-10. The red curve shows the result of training with one of the new learning rate policies. 
}
\label{fig:triangularCIFAR}
  \vspace{-5pt}
\end{figure}


Conventional wisdom dictates that the learning rate should be a single value that monotonically decreases during training.
This paper demonstrates the surprising phenomenon that a varying learning rate during training is beneficial overall and thus proposes to let the global learning rate vary cyclically within a band of values instead of setting it to a fixed value.
In addition, this cyclical learning rate (CLR) method practically eliminates the need to tune the learning rate yet achieve near optimal  classification accuracy.
Furthermore, unlike adaptive learning rates, the CLR methods require essentially no additional computation.

The potential benefits of CLR can be seen in Figure \ref{fig:triangularCIFAR}, which shows the test data classification accuracy of the CIFAR-10 dataset during training\footnote{Hyper-parameters and architecture were obtained in April 2015 from caffe.berkeleyvision.org/gathered/examples/cifar10.html}.
The baseline (blue curve) reaches a final accuracy of $81.4\%$ after $70,000$ iterations.
In contrast, it is possible to fully train the network using the CLR method instead of tuning (red curve) within 25,000 iterations and attain the same accuracy.

The contributions of this paper are:
\begin{enumerate}
\item A methodology for setting the global learning rates for training neural networks that eliminates the need to perform numerous experiments to find the best values and schedule with essentially no additional computation.
\item A surprising phenomenon is demonstrated - allowing the learning rate to rise and fall is beneficial overall even though it might temporarily harm the network's performance.
\item Cyclical learning rates are demonstrated with ResNets, Stochastic Depth networks, and DenseNets on the CIFAR-10 and CIFAR-100 datasets, and on ImageNet with two well-known architectures: AlexNet \cite{Alexnet12} and GoogleNet \cite{szegedy2014going}.
\end{enumerate}

\section{Related work}

The book ``Neural Networks: Tricks of the Trade'' is a terrific source of practical advice.
In particular, Yoshua Bengio \cite{Bengio12} discusses reasonable ranges for learning rates and stresses the importance of tuning the learning rate.
A technical report by Breuel \cite{breuel2015effects} provides guidance on a variety of hyper-parameters.
There are also a numerous websites giving practical suggestions for setting the learning rates.

\textbf{Adaptive learning rates}:
Adaptive learning rates can be considered a competitor to cyclical learning rates because one can rely on local adaptive learning rates in place of global learning rate experimentation but there is a significant computational cost in doing so.
CLR does not possess this computational costs so it can be used freely.

A review of the early work on adaptive learning rates can be found in George and Powell \cite{george06}.
Duchi, \etal \cite{duchi11} proposed AdaGrad, which is one of the early adaptive methods that estimates the learning rates from the gradients.

RMSProp is discussed in the slides by Geoffrey Hinton\footnote{www.cs.toronto.edu/~tijmen/csc321/slides/lecture\_slides\_lec6.pdf} \cite{tieleman2012lecture}.
RMSProp is described there as ``Divide the learning rate for a weight by a running average of the magnitudes of recent gradients for that weight.''
RMSProp is a fundamental adaptive learning rate method that others have built on.

Schaul \etal \cite{PeskyLR12} discuss an adaptive learning rate based on  a  diagonal estimation of the Hessian of the gradients.
One of the  features of their method is that they allow their automatic method to decrease or increase the learning rate.
However, their paper seems to limit the idea of increasing learning rate to non-stationary problems.  
On the other hand, this paper demonstrates that a schedule of increasing the learning rate is  more universally valuable. 

Zeiler \cite{zeiler2012adadelta}  describes his AdaDelta method, which improves on AdaGrad based on two ideas: limiting the sum of squared gradients over all time to a limited window, and making the parameter update rule consistent with a units evaluation on the relationship between the update and the Hessian.

More recently, several papers have appeared on adaptive learning rates.
Gulcehre and Bengio \cite{gulcehre2014adasecant}  propose an adaptive learning rate algorithm, called AdaSecant, that utilizes the root mean square statistics and variance of the gradients.
Dauphin \etal \cite{dauphin2015rmsprop} show that RMSProp provides a biased estimate and go on to describe another estimator, named ESGD, that is unbiased.
Kingma and Lei-Ba \cite{Adam15}  introduce Adam that is designed to combine the advantages from AdaGrad and RMSProp.
Bache, \etal \cite{bache2014hot} propose exploiting solutions to a multi-armed bandit problem for learning rate selection.
A summary and tutorial of adaptive learning rates can be found in a recent paper by Ruder \cite{ruder2016overview}.

Adaptive learning rates are fundamentally different from CLR policies, and CLR can be combined with adaptive learning rates, as shown in Section \ref{sec:cifar}.  
In addition, CLR policies are computationally simpler than adaptive learning rates.
CLR is likely most similar to the SGDR method \cite{loshchilov2016sgdr} that appeared recently.

\section{Optimal Learning Rates}

\subsection{Cyclical Learning Rates}
\label{sec:CLR}

\begin{figure}[htb]
\begin{center}
  \vspace{-10pt}
   \includegraphics[width=0.8\linewidth]{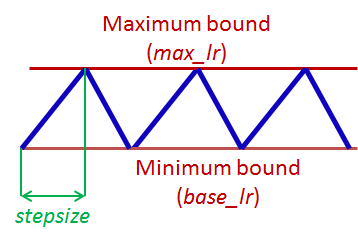}
\end{center}
  \vspace{-15pt}
   \caption{Triangular learning rate policy.  The blue lines represent learning rate values changing  between bounds.  
The input parameter $stepsize$ is the number of iterations in half a cycle.
}
\label{fig:triangularWindow}
 \vspace{-5pt}
\end{figure}

The essence of this learning rate policy comes from the observation that increasing the learning rate might have a short term negative effect and yet achieve a longer term beneficial effect.
This observation leads to the idea of letting the learning rate vary within a range of values rather than adopting a stepwise fixed or exponentially decreasing value.
That is, one sets minimum and maximum boundaries and the learning rate cyclically varies between these bounds.
Experiments with numerous functional forms, such as a triangular window (linear), a Welch window (parabolic) and a Hann window (sinusoidal) all produced equivalent results 
This led to adopting a triangular window (linearly increasing then linearly decreasing), which is illustrated in Figure \ref{fig:triangularWindow}, because it is the simplest function that incorporates this idea.
The rest of this paper refers to this as the $triangular$ learning rate policy.

An intuitive understanding of why CLR methods work comes from considering the loss function topology.  
Dauphin \etal \cite{dauphin2015rmsprop} argue that the difficulty in minimizing the loss arises from saddle points rather than poor local minima.
Saddle points have small gradients that slow the learning process. 
However, increasing the learning rate allows more rapid traversal of saddle point plateaus.
A more practical reason as to why CLR works is that, by following the methods in Section \ref{sec:bounds}, it is likely the optimum learning rate will be between the bounds and near optimal learning rates will be used throughout training.

The red curve in Figure \ref{fig:triangularCIFAR} shows the result of the $triangular$ policy on CIFAR-10.
The settings used to create the red curve were a minimum learning rate of $0.001$ (as in the original parameter file) and a maximum of $0.006$.
Also, the cycle length (i.e., the number of iterations until the learning rate returns to the initial value) is set to $4,000$ iterations (i.e., $stepsize = 2000$) and  Figure \ref{fig:triangularCIFAR} shows that the accuracy peaks at the end of each cycle.

Implementation of the code for a new learning rate policy is straightforward.
An example of the code added to Torch 7 in the experiments shown in Section \ref{sec:cifarResNet} is the following few lines: 
\lstset{language=C}
\lstset{basicstyle=\footnotesize}
\begin{lstlisting}
   local cycle = math.floor(1 + 
       epochCounter/(2*stepsize))
   local x = math.abs(epochCounter/stepsize 
       - 2*cycle + 1)
   local lr = opt.LR + (maxLR - opt.LR) 
       * math.max(0,  (1-x))
\end{lstlisting}
where $opt.LR $ is the specified lower (i.e., base) learning rate, $epochCounter$ is the number of epochs of training, and $lr$ is the computed learning rate.
This policy is named $triangular$ and is as described above, with two new input parameters defined: $stepsize$  (half the period or cycle length) and $max\_lr$ (the maximum learning rate boundary).
This code  varies the learning rate linearly between the minimum ($base\_lr$) and the maximum ($max\_lr$).



In addition to the triangular policy, the following CLR policies are discussed in this paper:
\begin{enumerate}
\item $triangular2$; the same as the $triangular$ policy except the learning rate difference is cut in half at the end of each cycle.  
This means the learning rate difference drops after each cycle.
\item $exp\_range$; the learning rate varies between the minimum and maximum boundaries and each boundary value declines by an exponential factor of $gamma^{iteration}$.
\end{enumerate}

\subsection{How can one estimate a good value for the cycle length?}
\label{sec:stepsize}

The length of a cycle and  the input parameter $stepsize$ can be easily computed from the number of iterations in an epoch.
An epoch is calculated by dividing the number of training images by the $batchsize$ used.
For example, CIFAR-10 has $50,000$ training images and the $batchsize$ is $100$ so an epoch $= 50,000 / 100 = 500$ iterations.
The final accuracy results are actually quite robust to cycle length but 
experiments show that it often is good to set $stepsize$ equal to $2 - 10$ times the number of iterations in an epoch.
For example, setting $stepsize = 8*epoch$ with the CIFAR-10 training run (as shown in Figure \ref{fig:triangularCIFAR}) only gives slightly better results than setting $stepsize = 2*epoch$. 

Furthermore, there is a certain elegance to the rhythm of these cycles and it simplifies the decision of when to drop learning rates and when to stop the current training run.
Experiments show that replacing each step of a constant learning rate with at least 3 cycles trains the network weights most of the way and  running for 4 or more cycles will achieve even better performance.
Also, it is best to stop training at the end of a cycle, which is when the learning rate is at the minimum value and the accuracy peaks.


\subsection{How can one estimate reasonable minimum and maximum boundary values?}
\label{sec:bounds}

There is a simple way to estimate reasonable minimum and maximum boundary values with  one  training run of the network for a few epochs.
It is a ``LR range test''; run your model for several epochs while letting the learning rate increase linearly between low and high LR values.
This test is enormously valuable whenever you are facing a new architecture or dataset.


\begin{figure}[htb]
  \vspace{-5pt}
  \begin{center}
   \includegraphics[width=1.0\linewidth]{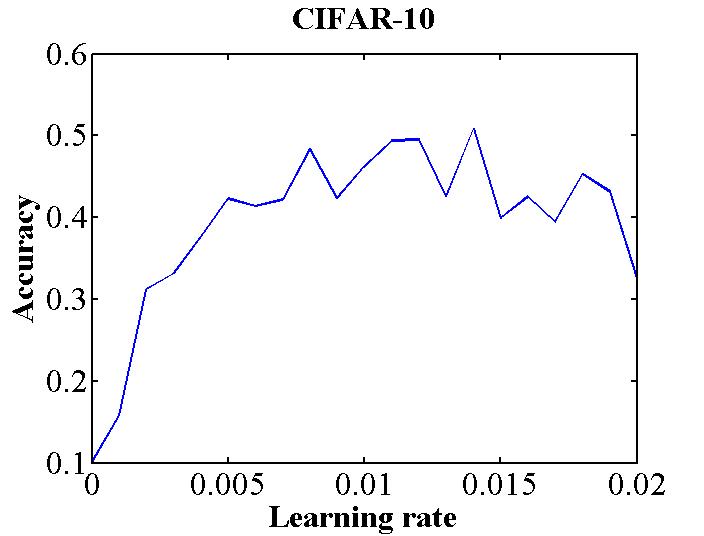}
  \end{center}
  \vspace{-5pt}
   \caption{Classification accuracy as a function of increasing learning rate for 8 epochs (LR range test). 
}
\label{fig:InitRangeCIFAR}
\end{figure}


The $triangular$ learning rate policy provides a simple mechanism to do this.
For example, in Caffe, set $base\_lr$ to the minimum value and set $max\_lr$ to the maximum value.
Set both the $stepsize$ and $max\_iter$ to the same number of iterations.
In this case, the learning rate will increase linearly from the minimum value to the maximum value during this short run.
Next, plot the accuracy versus learning rate. 
Note the learning rate value when the accuracy starts to increase and when the accuracy slows, becomes ragged, or starts to fall.
These two learning rates are good choices for bounds; that is, set $base\_lr$ to the first value and set $max\_lr$ to the latter value.
Alternatively, one can use the rule of thumb that the optimum learning rate is usually within a factor of two of the largest one that converges \cite{Bengio12} and set $base\_lr$ to $\frac{1}{3}$ or $\frac{1}{4}$ of $max\_lr$.

Figure \ref{fig:InitRangeCIFAR} shows an example of making this type of run with the CIFAR-10 dataset, using the architecture and hyper-parameters provided by Caffe.
One can see from Figure \ref{fig:InitRangeCIFAR} that the model  starts converging right away, so it is reasonable to set $ base\_lr = 0.001 $.
Furthermore, above a learning rate of $ 0.006 $  the accuracy rise gets rough and eventually begins to drop so it is reasonable to set $max\_lr = 0.006$.

Whenever one is starting with a new architecture or dataset, a single LR range test provides both a good LR value and a good range.
Then one should compare runs with a fixed LR versus CLR with this range.
Whichever wins can be used with confidence for the rest of one's experiments.


%
%


\begin{table}[tb]
\begin{center}
  \begin{tabular}{| c | c | c | c |}
    \hline
   Dataset  & LR policy & Iterations & Accuracy (\%) \\ \hline \hline
   CIFAR-10 &  $ fixed $  & 70,000 & 81.4 \\ \hline
   CIFAR-10 &  $ triangular2 $  & $\mathbf{25,000}$ & 81.4 \\ \hline
   CIFAR-10 &  $ decay $  & 25,000 &  78.5 \\ \hline
   CIFAR-10 &  $  exp $  & 70,000 & 79.1 \\ \hline
   CIFAR-10 &  $  exp\_range $  & 42,000 & $\mathbf{82.2}$ \\ \hline  \hline
   AlexNet &  $  fixed $  & 400,000 & 58.0 \\ \hline
   AlexNet &  $  triangular2 $  & 400,000 & $\mathbf{58.4}$ \\ \hline
   AlexNet &  $  exp $  & 300,000 & 56.0 \\ \hline
   AlexNet &  $  exp $  & 460,000 & 56.5 \\ \hline
   AlexNet &  $  exp\_range $  & 300,000 & 56.5 \\ \hline  \hline
   GoogLeNet &  $  fixed $  & 420,000 & 63.0 \\ \hline
   GoogLeNet &  $  triangular2 $  & 420,000 & $\mathbf{64.4}$ \\ \hline
   GoogLeNet &  $  exp  $ & 240,000 & 58.2 \\ \hline
   GoogLeNet &  $  exp\_range $  & 240,000 & 60.2 \\ \hline
  \end{tabular}
  \vspace{5pt}
  \caption{Comparison of accuracy results on test/validation data at the end of the training.}
  \label{tab:results}
\end{center}
  \vspace{-10pt}
\end{table}


\section{Experiments}
\label{sec:examples}

The purpose of this section is to demonstrate the effectiveness of the CLR methods on some standard datasets and with a range of architectures.
In the subsections below, CLR policies are used for training with the CIFAR-10, CIFAR-100, and ImageNet datasets.
These three datasets and a variety of architectures demonstrate the versatility of CLR.

\begin{figure}[tb]
  \vspace{-5pt}
  \begin{center}
   \includegraphics[width=1.0\linewidth]{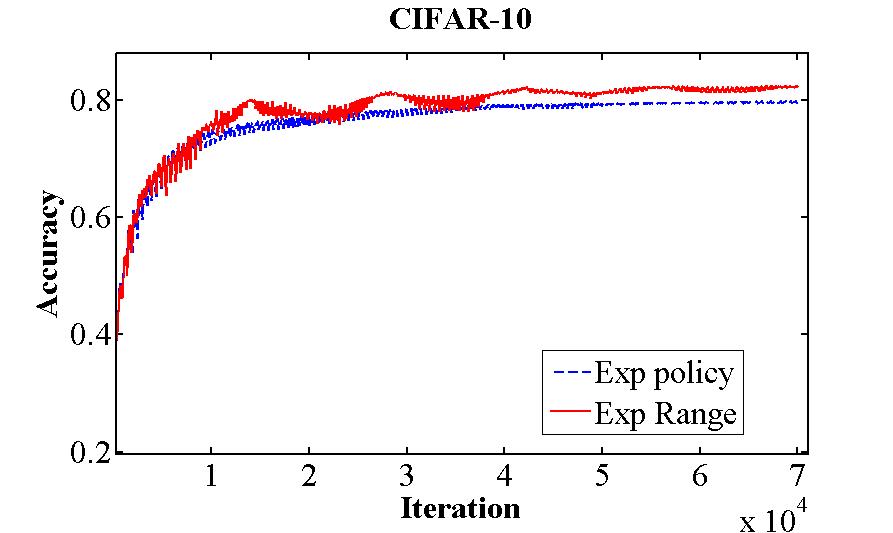}
  \end{center}
  \vspace{-5pt}
   \caption{Classification accuracy as a function of iteration for $70,000$ iterations. }
\label{fig:cifar_exp_range}
\end{figure}

\begin{figure}[tb]
  \begin{center}
   \includegraphics[width=1.0\linewidth]{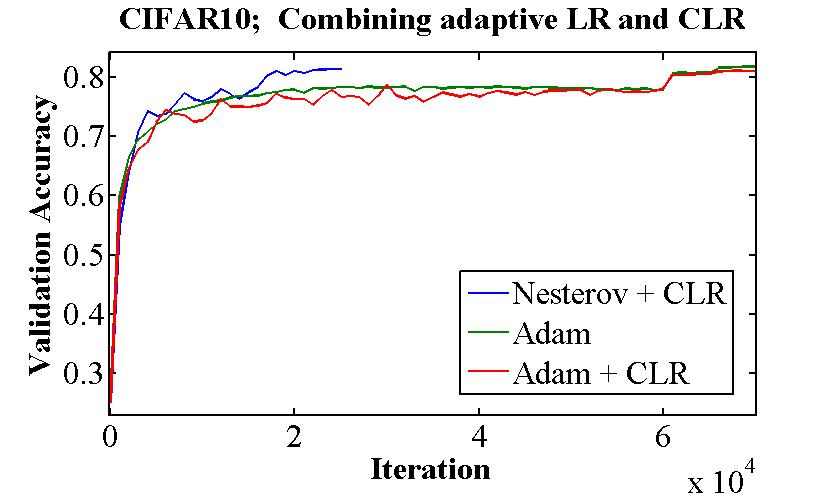}
  \end{center}
  \vspace{-5pt}
   \caption{Classification accuracy as a function of iteration for the CIFAR-10 dataset using adaptive learning methods. See text for explanation. }
\label{fig:cifar_adaptiveLR}
  \vspace{-5pt}
\end{figure}

\subsection{CIFAR-10 and CIFAR-100}
\label{sec:cifar}

\subsubsection{Caffe's CIFAR-10 architecture}
\label{sec:cifarCaffe}

The  CIFAR-10 architecture and hyper-parameter settings on the Caffe website are fairly standard and were used here  as a baseline.
As discussed in Section \ref{sec:stepsize}, an epoch is equal to $500$ iterations and a good setting for $stepsize$ is $2,000$.
Section \ref{sec:bounds} discussed how to estimate reasonable minimum and maximum boundary values for the learning rate from Figure \ref{fig:InitRangeCIFAR}.
All that is needed to optimally train the network is to set $base\_lr = 0.001$ and  $max\_lr = 0.006$.
This is all that is needed to optimally train the network.
For the $triangular2$ policy run shown in Figure \ref{fig:triangularCIFAR}, the $stepsize$ and learning rate bounds are shown in Table \ref{tab:cifarParams}.


\begin{table}[htb]
\begin{center}
  \begin{tabular}{| c | c | c | c | c |}
    \hline
    base\_lr & max\_lr & stepsize & start & max\_iter \\ \hline \hline
    0.001 & 0.005 &   2,000 & 0 & 16,000 \\ \hline
    0.0001 & 0.0005 &  1,000 & 16,000 & 22,000 \\ \hline
    0.00001 & 0.00005  & 500 & 22,000 & 25,000 \\ \hline
  \end{tabular}
  \vspace{5pt}
  \caption{Hyper-parameter settings for CIFAR-10 example in Figure \ref{fig:triangularCIFAR}.}
  \label{tab:cifarParams}
\end{center}
  \vspace{-5pt}
\end{table}


Figure \ref{fig:triangularCIFAR} shows the result of running with the $triangular2$ policy with the parameter setting in Table \ref{tab:cifarParams}.
As shown in Table \ref{tab:results}, one obtains the same test classification accuracy of $81.4\%$ after only $25,000$ iterations with the $triangular2$ policy as obtained by running the standard hyper-parameter settings for $70,000$ iterations.  

One might speculate that the benefits from the $triangular$ policy derive from reducing the learning rate because this is when the accuracy climbs the most.
As a test, a $decay$ policy was implemented where the learning rate starts at the $max\_lr$ value and then is linearly reduced to the $base\_lr$ value for $stepsize$ number of iterations.
After that, the learning rate is fixed to $base\_lr$.
For the $decay$ policy, $max\_lr = 0.007$, $base\_lr = 0.001$,  and $stepsize = 4000$.
Table \ref{tab:results} shows that the final accuracy is only $78.5\%$, providing evidence that  \textbf{both increasing and decreasing} the learning rate are essential for the benefits of the CLR method.

Figure \ref{fig:cifar_exp_range} compares the $exp$ learning rate policy in Caffe with the new $exp\_range$ policy using $gamma = 0.99994$ for both policies.
The result is that when using the $exp\_range$ policy one can stop training at iteration $42,000$ with a test accuracy of $82.2\%$ (going to iteration $70,000$ does not improve on this result).
This is substantially better than the best test accuracy of $79.1\%$ one obtains from using the $exp$ learning rate policy.


\begin{figure}[tb]
  \begin{center}
   \includegraphics[width=1.0\linewidth]{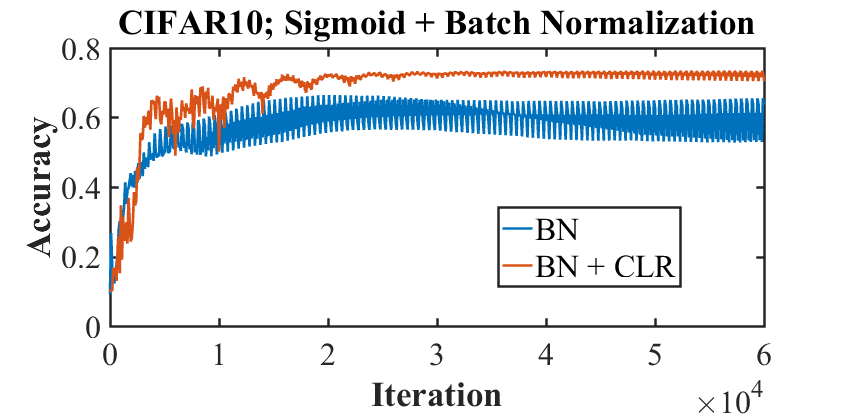}
  \end{center}
  \vspace{-5pt}
   \caption{Batch Normalization CIFAR-10 example (provided with the Caffe download).  }
\label{fig:bnclr}
  \vspace{-5pt}
\end{figure}


The current Caffe download contains additional architectures and hyper-parameters for CIFAR-10 and in particular there is one with sigmoid non-linearities and batch normalization.
Figure \ref{fig:bnclr} compares the training accuracy  using the downloaded hyper-parameters with a fixed learning rate (blue curve) to using a cyclical learning rate (red curve).
As can be seen in this Figure, the final accuracy for the fixed learning rate (60.8\%) is substantially lower than the cyclical learning rate final accuracy (72.2\%).
There is clear performance improvement when using CLR with this architecture containing sigmoids and batch normalization.

\begin{table}[htb]
\begin{center}
  \begin{tabular}{| c | c | c | c |}
    \hline
   LR type/bounds  & LR policy & Iterations & Accuracy (\%) \\ \hline \hline
   Nesterov \cite{nesterov1983method} &  $ fixed $  & 70,000 & 82.1 \\ \hline
   0.001 - 0.006  &  $ triangular $  & 25,000 & 81.3 \\ \hline
   ADAM \cite{Adam15} &  $ fixed $  & 70,000 & 81.4 \\ \hline
   0.0005 - 0.002  &  $ triangular $  & 25,000 & 79.8 \\ \hline
     &  $ triangular $  & 70,000 & 81.1 \\ \hline
   RMSprop \cite{tieleman2012lecture} &  $ fixed $  & 70,000 & 75.2 \\ \hline
   0.0001 - 0.0003 &  $ triangular $  & 25,000 & 72.8  \\ \hline
   &  $ triangular $  & 70,000 & 75.1  \\ \hline
   AdaGrad \cite{duchi11} &  $ fixed $  & 70,000 &  74.6 \\ \hline
   0.003 - 0.035  &  $ triangular $  & 25,000 & 76.0  \\ \hline
   AdaDelta \cite{zeiler2012adadelta} &  $ fixed $  & 70,000 & 67.3 \\ \hline
   0.01 - 0.1  &  $ triangular $  & 25,000 &  67.3 \\ \hline
  \end{tabular}
  \vspace{5pt}
  \caption{Comparison of CLR with adaptive learning rate methods. The table shows accuracy results for the CIFAR-10 dataset on test data at the end of the training.}
  \label{tab:adaptiveLR}
\end{center}
  \vspace{-5pt}
\end{table}


Experiments were carried out with architectures featuring both adaptive learning rate methods and CLR.
Table \ref{tab:adaptiveLR} lists the final accuracy values from various adaptive learning rate methods, run with and without CLR.
All of the adaptive methods in Table \ref{tab:adaptiveLR} were run by invoking the respective option in Caffe.
The learning rate boundaries are given in Table \ref{tab:adaptiveLR} (just below the method's name), which were determined by using the technique described in Section \ref{sec:bounds}. Just the lower bound was used for $ base\_lr $ for the $ fixed $ policy.

Table \ref{tab:adaptiveLR} shows that for some adaptive learning rate methods combined with CLR, the final accuracy after only 25,000 iterations is equivalent to the accuracy obtained without CLR after 70,000 iterations.  
For others, it was necessary (even with CLR) to run until 70,000 iterations to obtain similar results.
Figure \ref{fig:cifar_adaptiveLR} shows the curves from running the Nesterov method with CLR (reached 81.3\% accuracy in only 25,000 iterations) and the Adam method both with and without CLR (both needed 70,000 iterations).
When using adaptive learning rate methods, the benefits from CLR are sometimes reduced, but CLR can still valuable as it sometimes provides benefit at essentially no cost.


\subsubsection{ResNets, Stochastic Depth, and DenseNets}
\label{sec:cifarResNet}

Residual networks \cite{he2015deep, he2016identity}, and the  family of variations that have subsequently emerged, achieve \SotA results on a variety of tasks.
Here we provide comparison experiments between the original implementations and versions with CLR for three members of this residual network family: the original ResNet \cite{he2015deep}, Stochastic Depth networks \cite{huang2016deep}, and the recent DenseNets \cite{huang2016densely}.
Our experiments can be readily replicated because the authors of these papers make their Torch code available\footnote{https://github.com/facebook/fb.resnet.torch, https://github.com/yueatsprograms/Stochastic\_Depth, https://github.com/liuzhuang13/DenseNet}.
Since all three implementation are available using the Torch 7 framework, the experiments in this section were performed using Torch.
In addition to the experiment in the previous Section, these networks also incorporate batch normalization \cite{ioffe2015batch} and demonstrate the value of CLR for architectures with batch normalization.

Both CIFAR-10 and the CIFAR-100 datasets were used in these experiments.
The CIFAR-100 dataset is similar to the CIFAR-10 data but it has 100 classes instead of 10 and each class has 600 labeled examples.

\begin{table}[htb]
\begin{center}
  \begin{tabular}{| c | c | c | }
    \hline
   Architecture  & CIFAR-10 (LR) & CIFAR-100 (LR) \\ \hline \hline
   ResNet   &  $ 92.8 (0.1) $  & $ 71.2 (0.1) $ \\ \hline
   ResNet  &  $  93.3 (0.2)$  & $  71.6 (0.2)$ \\ \hline
   ResNet  &  $ 91.8 (0.3)$  & $  71.9 (0.3)$ \\ \hline
   ResNet+CLR   &  $ 93.6 (0.1-0.3) $  & $ 72.5 (0.1-0.3)$ \\ \hline
   SD &  $ 94.6 (0.1) $  & $ 75.2 (0.1)$ \\ \hline
   SD &  $ 94.5 (0.2) $  & $ 75.2 (0.2)$ \\ \hline
   SD &  $ 94.2 (0.3 )$  & $ 74.6 (0.3)$ \\ \hline
   SD+CLR &  $ 94.5 (0.1-0.3)$  & $ 75.4 (0.1-0.3)$ \\ \hline
   DenseNet  &  $ 94.5 (0.1)$  & $ 75.2 (0.1)$ \\ \hline
   DenseNet  &  $ 94.5 (0.2)$  & $ 75.3 (0.2)$ \\ \hline
   DenseNet  &  $ 94.2 (0.3)$  & $ 74.5 (0.3)$ \\ \hline
   DenseNet+CLR  &  $ 94.9 (0.1-0.2)$  & $ 75.9  (0.1-0.2)$ \\ \hline
  \end{tabular}
  \vspace{5pt}
  \caption{Comparison of CLR with ResNets \cite{he2015deep, he2016identity}, Stochastic Depth (SD) \cite{huang2016deep}, and DenseNets \cite{huang2016densely}. The table shows the average accuracy of 5 runs for the CIFAR-10 and CIFAR-100 datasets on test data at the end of the training.}
  \label{tab:ResNets}
\end{center}
 \vspace{-15pt}
\end{table}

The results for these two datasets on these three architectures are summarized in Table \ref{tab:ResNets}.
The left  column give the architecture and whether CLR was used in the experiments.
The other two columns gives the average final accuracy from five runs and the initial learning rate or range used in parenthesis, which are reduced (for both the fixed learning rate and the range) during the training according to the same schedule used in the original implementation.
For all three architectures, the original implementation uses an initial LR of 0.1 which we use as a baseline.

The accuracy results in Table \ref{tab:ResNets} in the right two columns are the average final test accuracies of five runs.
The Stochastic Depth implementation was slightly different than the ResNet and DenseNet implementation in that the authors split the 50,000 training images into  45,000 training images and 5,000 validation images.
However, the reported results in Table \ref{tab:ResNets} for the SD architecture is only test accuracies for the five runs.
The learning rate range used by CLR was determined by the LR range test method and the cycle length was choosen as a tenth of the maximum number of epochs that was specified in the original implementation.

In addition to the accuracy results shown in Table \ref{tab:ResNets}, similar results were obtained in Caffe for DenseNets \cite{huang2016densely} on CIFAR-10 using the prototxt files provided by the authors.
The average accuracy of five runs with learning rates of $ 0.1, 0.2, 0.3$ was $ 91.67\%, 92.17\%, 92.46\%, $ respectively, but running with CLR within the range of 0.1 to 0.3, the average accuracy was  $ 93.33\%$.

The results from all of these experiments show similar or better accuracy performance when using CLR versus using a fixed learning rate, even though the performance drops at some of the learning rate values within this range.
These experiments confirm that it is beneficial to use CLR for a variety of residual architectures and for both CIFAR-10 and CIFAR-100.

\begin{figure}[tb]
  \begin{center}
   \includegraphics[width=1.0\linewidth]{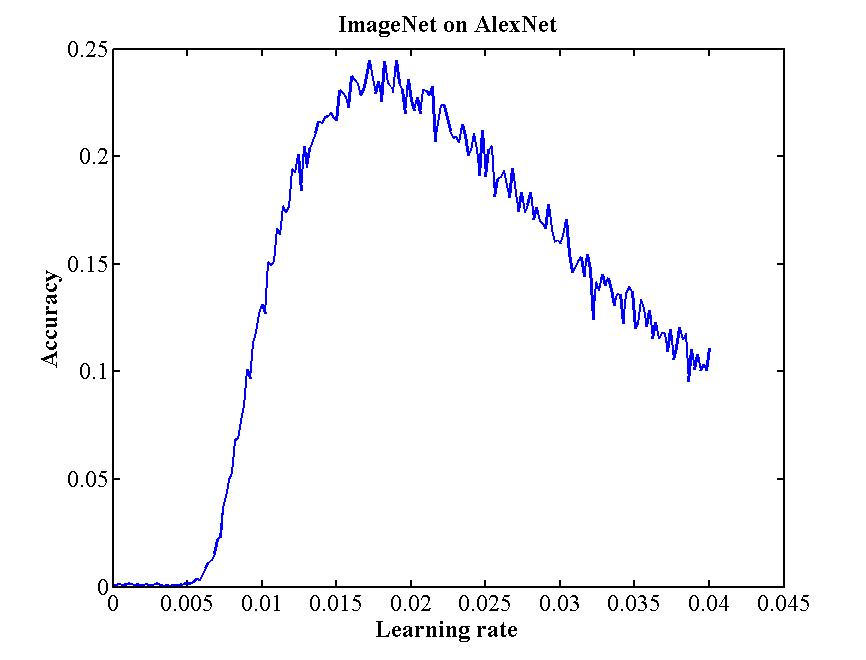}
  \end{center}
  \vspace{-5pt}
   \caption{AlexNet LR range test; validation classification accuracy as a function of increasing learning rate. }
\label{fig:InitRangeAlexNet}
  \vspace{-10pt}
\end{figure}

\begin{figure}[tb]
  \begin{center}
   \includegraphics[width=1.0\linewidth]{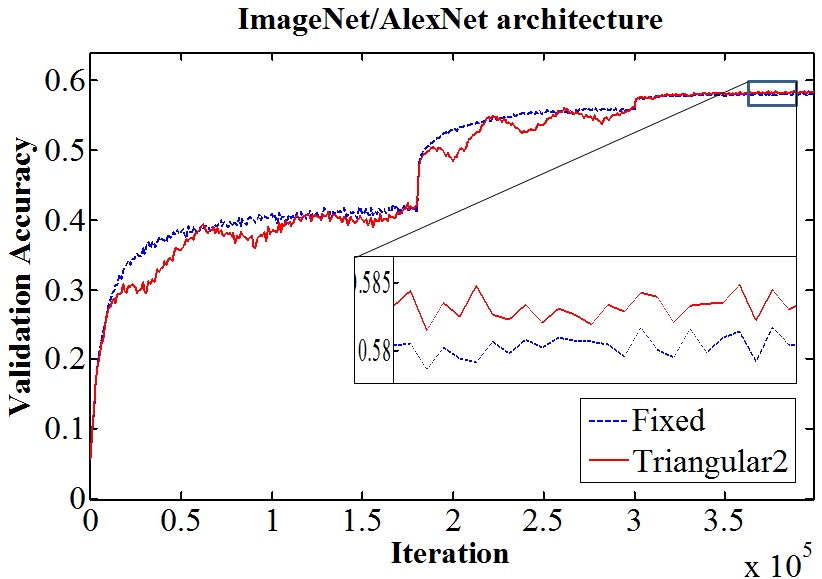}
  \end{center}
  \vspace{-5pt}
   \caption{Validation data classification accuracy as a function of iteration  for $fixed$ versus $triangular$.
}
\label{fig:alexTri}
  \vspace{-5pt}
\end{figure}

\subsection{ImageNet}

The ImageNet dataset \cite{Imagenet15} is often used in deep learning literature as a standard for comparison.
The ImageNet classification challenge provides about $1,000$  training images for each of the $1,000$ classes, giving  a total of $1,281,167$ labeled training images.

\subsubsection{AlexNet}
\label{sec:alexnet}

The Caffe website provides the architecture and hyper-parameter files for a slightly modified AlexNet  \cite{Alexnet12}. 
These were downloaded from the website and used as a baseline.
In the training results reported in this section, all weights were initialized the same so as to avoid differences due to different random initializations.

Since the $batchsize$ in the architecture file is $256$, an epoch is equal to $1,281,167 / 256 = 5,005$ iterations.
Hence, a reasonable setting for $stepsize$ is $6$ epochs or $30,000$ iterations.

Next, one can estimate reasonable minimum and maximum boundaries for the learning rate  from Figure \ref{fig:InitRangeAlexNet}.
It can be seen from this figure that the training doesn't start converging until at least $0.006$ so setting $base\_lr = 0.006$ is reasonable.
However, for a fair comparison to the baseline where $base\_lr = 0.01$, it is necessary to set the $base\_lr$ to $0.01$ for the $triangular$ and $triangular2$ policies or else the majority of the apparent improvement in the accuracy will be from the smaller learning rate.
As for the maximum boundary value, the training peaks and drops above a learning rate of $0.015$ so $max\_lr = 0.015$ is reasonable.
For comparing the $exp\_range$ policy to the $exp$ policy, setting $base\_lr = 0.006$ and $max\_lr = 0.014$ is reasonable and in this case one expects that the average accuracy of the $exp\_range$ policy to be equal to the accuracy from the $exp$ policy.


\begin{figure}[tb]
  \begin{center}
   \includegraphics[width=1.0\linewidth]{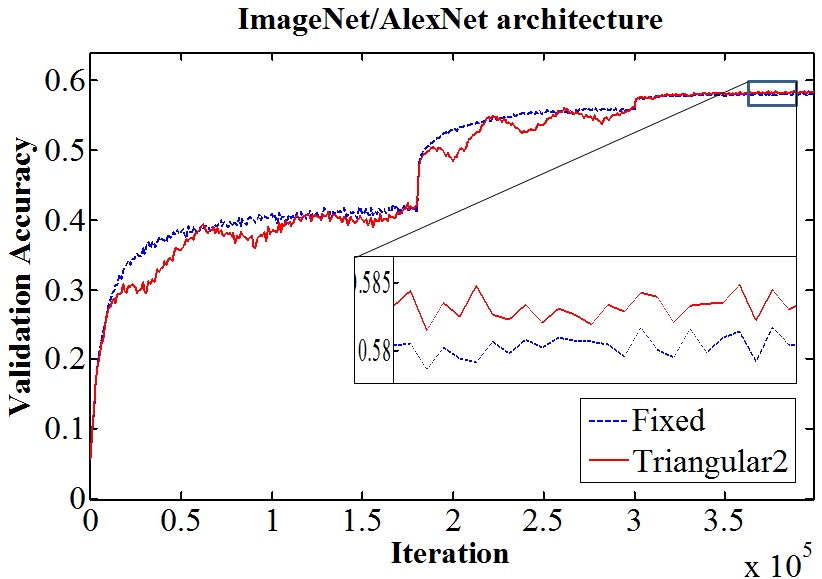}
  \end{center}
  \vspace{-5pt}
   \caption{Validation data classification accuracy as a function of iteration  for $fixed$ versus $triangular$.
}
\label{fig:alexTri}
  \vspace{-5pt}
\end{figure}


\begin{figure}[tb]
  \begin{center}
   \includegraphics[width=1.0\linewidth]{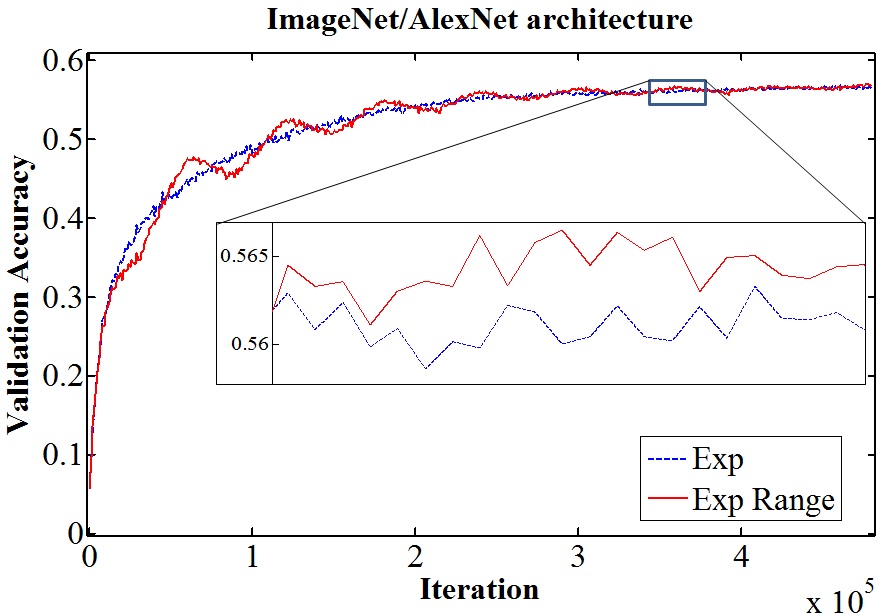}
  \end{center}
  \vspace{-5pt}
   \caption{Validation data classification accuracy as a function of iteration  for $exp$ versus $exp\_range$.
}
\label{fig:alexnet_exp}
  \vspace{-5pt}
\end{figure}


Figure \ref{fig:alexTri} compares the results of running with the $fixed$ versus the $triangular2$ policy for the AlexNet architecture.
Here, the peaks at iterations that are multiples of 60,000 should produce a classification accuracy that corresponds to the $fixed$ policy.
Indeed, the accuracy peaks at the end of a cycle for the $triangular2$ policy are similar to the accuracies from the standard $fixed$ policy, which implies that the baseline learning rates are set quite well (this is also implied by Figure \ref{fig:InitRangeAlexNet}).
As shown in Table \ref{tab:results}, the final accuracies from the CLR training run are only $0.4\%$ better than the accuracies from the $fixed$ policy.



Figure \ref{fig:alexnet_exp} compares the results of running with the $exp$ versus the $exp\_range$ policy for the AlexNet architecture with $gamma = 0.999995$ for both policies.
As expected, Figure \ref{fig:alexnet_exp} shows that the accuracies from the $exp\_range$ policy do oscillate around the $exp$ policy accuracies.
The advantage of the $exp\_range$ policy is that the accuracy of $56.5\%$ is already obtained at iteration $300,000$ whereas the $exp$ policy takes until iteration $460,000$ to reach $56.5\%$.

Finally, a comparison between the $fixed$ and $exp$ policies in Table \ref{tab:results} shows the $fixed$ and $triangular2$ policies produce accuracies that are almost $2\%$ better than their exponentially decreasing counterparts, but this difference is probably due to not having tuned $gamma$.


\begin{figure}[tb]
  \begin{center}
   \includegraphics[width=1.0\linewidth]{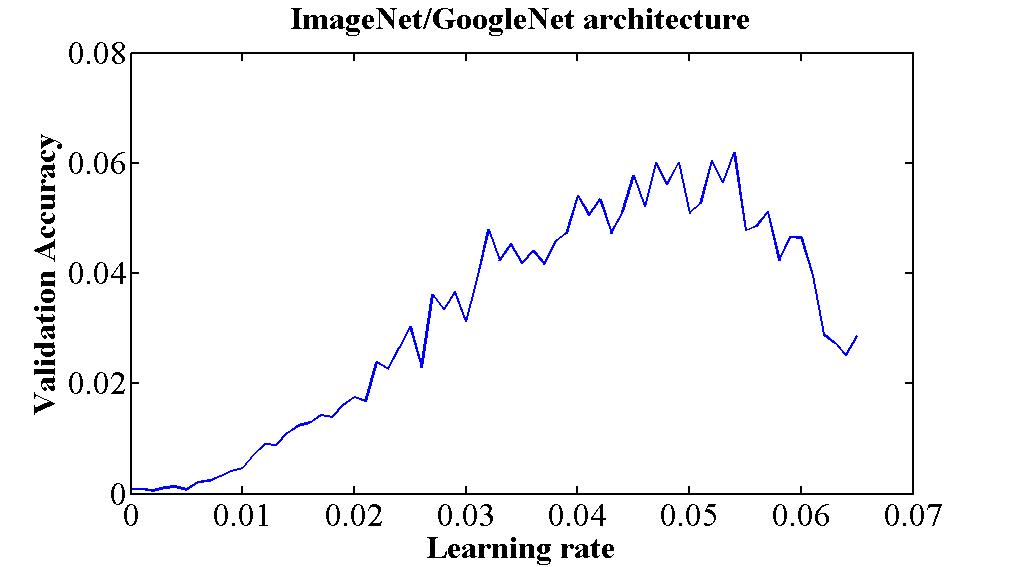}
  \end{center}
  \vspace{-5pt}
   \caption{GoogleNet LR range test; validation classification accuracy as a function of increasing learning rate. }
\label{fig:InitRangeGoogleNet}
  \vspace{-5pt}
\end{figure}


\subsubsection{GoogLeNet/Inception Architecture}

The GoogLeNet architecture was a winning entry to the ImageNet 2014 image classification competition.
Szegedy \etal \cite{szegedy2014going}  describe the architecture in detail but did not provide the architecture file.
The architecture file  publicly available from Princeton\footnote{vision.princeton.edu/pvt/GoogLeNet/} was used in the following experiments.
The GoogLeNet paper does not state the learning rate values and the  hyper-parameter solver file is not available for a baseline but not having these hyper-parameters is a typical situation when one is developing a new architecture or applying a network to a new dataset.
This is a situation that CLR readily handles. 
Instead of running numerous experiments to find optimal learning rates, the $base\_lr$ was set to a best guess value of $0.01$.

The first step is to estimate the $stepsize$ setting.
Since the architecture uses a batchsize of $128$ 
an epoch is equal to $1,281,167 / 128 = 10,009$ iterations.
Hence, good settings for $stepsize$ would be $20,000$, $30,000$, or possibly $40,000$.
The results in this section are based on $stepsize = 30000$.


\begin{figure}[tb]
  \begin{center}
   \includegraphics[width=1.0\linewidth]{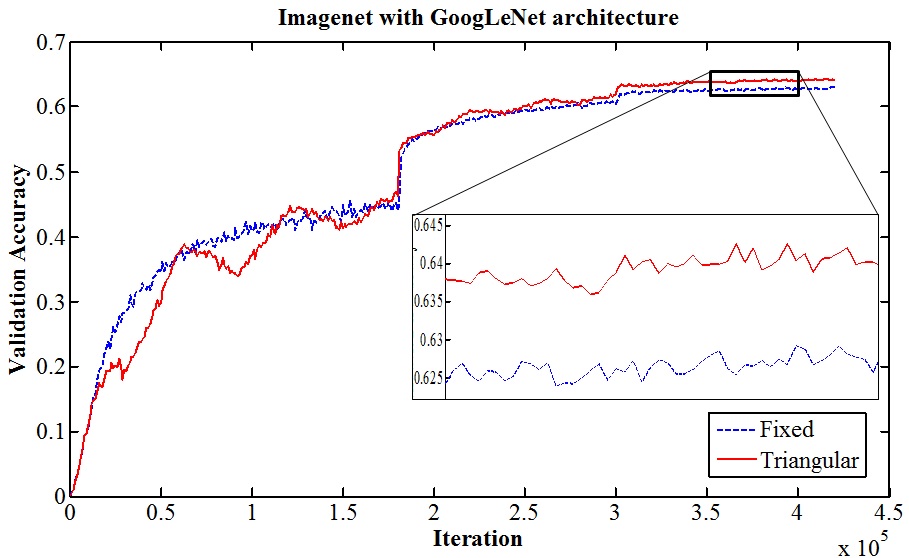}
\end{center}
  \vspace{-5pt}
   \caption{Validation data classification accuracy as a function of iteration for $fixed$ versus $triangular$.
}
\label{fig:googLeNet_fixed}
  \vspace{-5pt}
\end{figure}


%

The next step is to estimate the bounds for the learning rate, which is found with the LR range test by making a run for 4 epochs where the learning rate linearly increases from $ 0.001$ to $ 0.065$ (Figure \ref{fig:InitRangeGoogleNet}).
This figure shows that one can use bounds between $0.01$ and $0.04$ and still have the model reach convergence.
However, learning rates above $0.025$ cause the training to converge erratically.
For both $triangular2$ and the $exp\_range$ policies, the $base\_lr$ was set to $0.01$ and  $max\_lr$ was set to $0.026$.
As above, the accuracy peaks for both these learning rate policies correspond to the same learning rate value as the $fixed$ and $exp$ policies.
Hence, the comparisons below will focus on the peak accuracies from the LCR methods.

Figure \ref{fig:googLeNet_fixed} compares the results of running with the $fixed$ versus the $triangular2$ policy for this architecture (due to time limitations, each training stage was not run until it fully plateaued).
In this case, the peaks at the end of each cycle for the $triangular2$ policy produce better  accuracies than the $fixed$ policy.
The final accuracy shows an improvement from the network trained by the $triangular2$ policy (Table \ref{tab:results})  to be  $1.4\%$ better than the accuracy from the $fixed$ policy.
This demonstrates that the $triangular2$ policy improves on a ``best guess'' for a fixed learning rate.

\begin{figure}[tb]
  \begin{center}
   \includegraphics[width=1.0\linewidth]{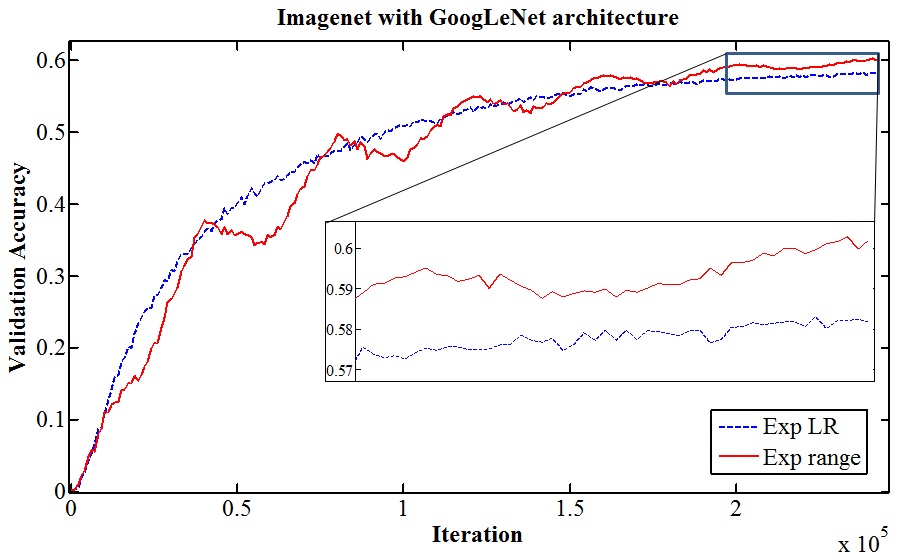}
\end{center}
  \vspace{-5pt}
   \caption{Validation data classification accuracy as a function of iteration  for $exp$ versus $exp\_range$.
	}
\label{fig:GoogleNet_exp}
  \vspace{-5pt}
\end{figure}

Figure \ref{fig:GoogleNet_exp} compares the results of running with the $exp$ versus the $exp\_range$ policy with $gamma = 0.99998$.
Once again, the peaks at the end of each cycle for the $exp\_range$ policy produce better validation accuracies than the $exp$ policy.
The final accuracy from the $exp\_range$ policy (Table \ref{tab:results}) is 2\% better than from the $exp$ policy.


\section{Conclusions}

The results presented  in this paper demonstrate the benefits of the cyclic learning rate (CLR) methods.
A short run of only a few epochs where the learning rate linearly increases is sufficient to estimate boundary learning rates for the CLR policies.
Then a policy where the learning rate cyclically varies between these bounds is sufficient to obtain near optimal classification results, often with fewer iterations.
This policy is easy to implement and unlike adaptive learning rate methods, incurs essentially no additional computational expense.

This paper shows that use of cyclic functions as a learning rate policy provides substantial improvements in performance for a range of architectures.
In addition, the cyclic nature of these methods provides guidance as to times to drop the learning rate  values (after 3 - 5 cycles) and when to stop the the training.
All of these factors reduce the guesswork in setting the learning rates and make these methods practical tools for everyone who trains neural networks.

This work has not explored the full range of applications for  cyclic learning rate methods.
We plan to determine if equivalent policies work for training different architectures, such as recurrent neural networks.
Furthermore, we believe that a theoretical analysis would provide an improved understanding of these methods, which might lead to improvements in the algorithms.


{\small
\bibliographystyle{ieee}
\bibliography{clr.bib}
}

\clearpage	

\appendix

\section{Instructions for adding CLR to Caffe}

Modify SGDSolver<Dtype>::GetLearningRate() which is in sgd\_solver.cpp (near line 38):

\lstset{language=C}
\lstset{basicstyle=\footnotesize}
\begin{lstlisting}
} else if (lr_policy == "triangular") {
  int itr = this->iter_ - this->param_.start_lr_policy();
  if(itr > 0) {
    int cycle = itr / (2*this->param_.stepsize());
    float x = (float) (itr - (2*cycle+1)*this->param_.stepsize());
    x = x / this->param_.stepsize();
    rate = this->param_.base_lr() + (this->param_.max_lr()- this->param_.base_lr()) 
           * std::max(double(0),(1.0 - fabs(x)));
  } else {
    rate = this->param_.base_lr();
  }
} else if (lr_policy == "triangular2") {
  int itr = this->iter_ - this->param_.start_lr_policy();
  if(itr > 0) {
    int cycle = itr / (2*this->param_.stepsize());
    float x = (float) (itr - (2*cycle+1)*this->param_.stepsize());
    x = x / this->param_.stepsize();
    rate = this->param_.base_lr() + (this->param_.max_lr()- this->param_.base_lr()) 
             * std::min(double(1), std::max(double(0), (1.0 -  fabs(x))/pow(2.0,double(cycle))));
  } else {
    rate = this->param_.base_lr();
}
\end{lstlisting}

Modify message SolverParameter which is in caffe.proto (near line 100):

\begin{lstlisting}
  optional float start_lr_policy = 41;
  optional float max_lr = 42; // The maximum learning rate for CLR policies
\end{lstlisting}

\section{Instructions for adding CLR to Keras}

Please see https://github.com/bckenstler/CLR.

\end{document}